# Towards human-like kinematics in industrial robotic arms: a case study on a UR3 robot

Adam Wolniakowski
*Faculty of Electrical Engineering*
*Bialystok University of Technology*
Bialystok, Poland
a..wolniakowski@pb.edu.pl

Kanstantsin Miatliuk
*Dept. of Robotics and Mechatronics*
*Bialystok University of Technology*
Bialystok, Poland
k.miatliuk@pb.edu.pl

Jose J. Quintana
*Dept. of Automatic and Electronic*
*University of Las Palmas de G.C.*
Las Palmas, Spain
josejuan.quintana@ulpgc.es

Miguel A. Ferrer
*IDeTIC Institute*
*University of Las Palmas de G.C.*
Las Palmas, Spain
miguelangel.ferrer@ulpgc.es

Moises Diaz
*IDeTIC Institute*
*University of Las Palmas de G.C.*
Las Palmas, Spain
moises.diaz@ulpgc.es

*Abstract*— Safety in industrial robotic environments is a hot research topic in the area of human-robot interaction (HRI). Up to now, a robotic arm on an assembly line interacts with other machines away from human workers. Nowadays, robotic arm manufactures are aimed to their robots could increasingly perform tasks collaborating with humans. One of the ways to improve this collaboration is by making the movement of robots more humanlike. This way, it would be easier for a human to foresee the movement of the robot and approach it without fear of contact. The main difference between the movement of a human and of a robotic arm is that the former has a bell-shaped speed profile while the latter has a uniform speed one. To generate this speed profile, the kinematic theory of rapid human movements and its Sigma-Lognormal model has been used. This model is widely used to explain most of the basic phenomena related to the control of human movements. Both human-like and robotic-like movements are transferred to the UR3 robot. In this paper we detail the how the UR3 robot was programmed to produce both kinds of movement. The dissimilarities result between the input motion and output motion to the robot confirm the possibility to develop human-like velocities in the UR3 robot.

## I. INTRODUCTION

Human-robot interaction (HRI) explores the improvements in cooperation between humans and machines. Traditionally it has been applied to service robots, but currently, it is becoming a fundamental pillar of Industry 4.0. In the recent past, the concept of an industrial robot was a robotic arm on an assembly line, interacting with other machines and away from humans. Things are changing and there are already many robots designed to work together with humans and the current trend of robotic arm manufacturers is that their robots can work together with humans to perform tasks together [1], it is what is known as collaborative robots or cobots [2].

One of the strategies followed in the HRI is to make the robots have anthropomorphic aspects, such as a structure of head, trunk, and limbs [3][4]. Other strategies are made in their relationship with humans based on speech, sounds, digital screens, ubiquitous systems, and also humanoid facial gestures [5]. Furthermore, robots are increasingly intended to be as autonomous as possible, with a great boom in artificial intelligence as support for the development of planning tasks or other more complex ones.

Among the different factors that intervene in the HRI, perhaps the least media, but not less important is the one related to the humanoid movement of robots. This movement is not only limited to the movement of the robot's limbs, but also to other movements such as the blinking of the eyes, the movement of the mouth, head, etc. This type of robotic movement is unconsciously perceived by humans and may pose a barrier or an approach to the uncanny valley for HRI [6][7].

In industrial environments, this rejection increases when it is necessary to work with robots in a collaborative way. One of the ways to improve these environments is by making the movement of robots more human. This way, it would be easier for a human to unconsciously intuit the movement of the robot [8][9].

One of the main differences between the movement of a human and a robotic arm is observed in the speed profiles of their movements. A robotic arm is generally programmed at a constant speed. In contrast, the speed profile of a human when executing a single movement is bell-shaped. Consequently, when the movement is long and complex, the shape of the human speed profile is an ordered sum of bells, one for each individual movement [10].

There are many theories and models that have attempted to describe the velocity profile of human movements. Some mathematical models provide analytical representations by exploiting the properties of various mathematical functions to reproduce human movements: exponentials, second-order systems, Gaussians, beta functions, splines, trigonometric functions, etc. Among the possible models, the kinematic theory of rapid human movements [11] and its Sigma-Lognormal model have been used to explain most of the basic phenomena reported in classical studies on the control of human movements [12] and to study various factors involved in fine motor skills [13]. Apart from these fundamental studies, the theory has been used, directly or indirectly, in many practical applications including the design of robotic trajectories [14].

The aim of this work is to show that the interaction of a human and a robotic arm will be friendlier the more the robotic movement resembles the human. For this, several robotic-like and human-like movements have been generated and implemented in an industrial robotic arm. The robotic movements have been generated with the basic programming commands of the robot and the humans have been synthesized based on the Sigma-Lognormal model.

The rest of the paper is organized as follows: First, an analysis of human movement will be made and the method used to generate this movement with the robot will be exposed, the robotic-like and the human-like movements will also be generated. Second, both types of movements will be implemented in a robotic arm. Next, the movements will be executed in the robot and compared with the designed movements. Finally, the conclusions will be presented.

## II. Synthetic Movement Generation

In this section, we describe the computational model used to produce movements. This unified model generates both spatiotemporal movements with uniform velocity and human-like velocities.

### A. Motion in industrial arm robot

An industrial robot is programmed to be used in manufacturing purposes. Typically, they operate in a controlled environment, which is secure for human and efficiency for the programmed tasks. Observing their movements, they continuously repeat a programmed trajectory in terms of target points, which depends on the application. Reaching these points depend on the programmed speed. There is a technological trade-off between rapid speed, efficiency and security. As a consequence of this programming, a uniform velocity profile can be observed, which is composed of a flat speed values between target points and short slopes at the beginning and the end of the target points.

Instead, when a person produces an automatized movement to reach different target points, his neuromuscular system produces a different pattern in his velocity profile. Previous studies have concluded that these patterns can be approximated as a sum of lognormal functions. This lognormality principle is the basis of the theory of rapid movements [15, 10], which has been used to model the end-effector of human arm when produce handwritten signatures [16], among other examples of rapid human movements.

Accordingly, towards the generation of friendlier robotic movements, we developed a unify computational model to generate artificially movements with uniform and human-like movements, giving as output a spatiotemporal tuple $(x, y, z, t)_{PC}$

### B. Human-like movement generation

Firstly, we define a set of $N$ random target points, $tp_j, \forall j = 1, ..., N$, in tridimensional rectangular parallelepiped. The objective is to design a spatiotemporal trajectory between two target points with human-like velocity. It is achieved by terms of the Sigma-Lognormal model. Each pair of consecutive target points correspond to an execution of the neuromuscular system, which is namely as stroke. The velocity of each stroke is designed synthetically as a lognormal function, the total movement being a weighted sum of delayed lognormals. The velocity of the movement can be mathematically written as follows:

$$\vec{v}(t) = \sum_{j=1}^{N} \vec{v_j}(t) = \sum_{j=1}^{N} D_j \begin{bmatrix} cos\phi_j(t) \\ sin\phi_j(t) \end{bmatrix} D_j \Lambda\left(t; t_{0_j}, \mu_j, \sigma_j^2\right) \quad (1)$$

Where $N$ denotes the number of target points or lognormal strokes, $t$ the temporal sequence, $t_{0_j}$ the time of stroke command occurrence, $D_j$ the amplitude of the stroke, $\mu_j$ the stroke delay, $\sigma_j$ the stroke response time and $\Lambda\left(t; t_{0_j}, \mu_j, \sigma_j^2\right)$ is defined as:

$$\Lambda\left(t; t_{0_j}, \mu_j, \sigma_j^2\right) = \frac{D_j}{\sigma_j \sqrt{2\pi}(t - t_{0_j})} e^{\left\{-\frac{\left[ln(t - t_{0_j}) - \mu_j\right]^2}{2\sigma_j^2}\right\}} \quad (2)$$

and $\phi_j(t)$ being the angular position:

$$\phi_j(t) = \theta_{s_j} + \frac{\theta_{e_j} - \theta_{s_j}}{2} \left[1 + erf\left(\frac{ln\left(t - t_{0_j}\right) - \mu_j}{\sigma_j \sqrt{2}}\right)\right] \quad (3)$$

As a result, the trajectory of each stroke, whose derivative is $\vec{v_j}(t)$, consists in an arc of circle between two target points, $tp_j$ and $tp_{j+1}$. Its start and end angle being denoted $\theta_{s_j}$ and $\theta_{e_j}$ respectively.

For the synthesis of this kind of movements, a lognormal is fitted in with each stroke. Let us assume a single stroke velocity profile given by $v_j(t)$. The values of $D_j$, $\mu_j$ and $\sigma_j^2$ are set from the following two hypotheses: Firstly, the margins for natural human handwriting given in [16]. Secondly, it was heuristically observed that most of the lognormals are centered when the Biosecure-SONOFF is represented by using ScriptStudio [10], i.e. the lognormal peaks approach the center of the strokes. Therefore, we configure our skewness close to zero but positive and the kurtosis around 3.

The calculation of these values is suggested as follows. From Eq. 1, the distance $s(t)$ traveled at time $t$ is obtained as:

$$s(t) = \int_{-\infty}^{\infty} v_j(t)dt = \frac{D_j}{2}\left(1 + \text{erf}\left(\frac{\ln(t - t_{0_j}) - \mu_j}{\sqrt{2}\sigma_j}\right)\right) \quad (4)$$

Then, let $l_s$ be the length of a stroke, i.e. the length of the arc between two consecutive minima. From Eq. (4) we deduce:

$$l_s = \frac{D_j}{2}\left(1 + \text{erf}\left(\frac{\ln(t - t_{0_j}) - \mu_j}{\sqrt{2}\sigma_j}\right)\right) \quad (5)$$

As erf(3) → 1, a possible solution of Eq. (5) is:

$$D_j = l_s \quad (6)$$

$$\mu_j = \ln(t - t_{0_j}) - 3\sqrt{2}\sigma_j \quad (7)$$

Furthermore, as the lognormals are centered in the middle of the stroke with a low positive skew, their maximum or mode, defined by $e^{\mu_j - \sigma_j^2}$, is around $t_{max_j} = t_{min_{j-1}} + 0.05$ with a slightly left skew. Therefore, it holds that:

$$t_{max_j} - t_{0_j} = e^{\mu_j - \sigma_j^2} \quad (8)$$

Combining Eq. (7) and Eq. (8) we obtain:

$$\sigma_j^2 + 3\sqrt{2}\sigma_j - \ln k = 0 \quad (9)$$

Thus, in the case of an isolated stroke of length $l_s$ and duration 0.1 sec, the simplified approach leads us to estimate $\sigma_j^2$ as the positive solution of the simple second order equation.

Since industrial robots operates in tridimensional environments, the mathematical formulation is extended to 2.5 dimensions. Having in mind the proposal described in [17], once the velocity of each stroke was designed in a bidimensional plane, it was adapted to the corresponding tridimensional location according its target points. Therefore, the straight line between two target points within the tridimensional rectangular parallelepiped is resampled with the artificial generated trajectory following the two dimensional procedure [10]. As a result, a tridimensional spatiotemporal movement was obtained in terms of $(x, y, z, t)_{PC}$.

*C. Robotic-like movement generation*

One on the motivation of this procedure is to implement a unique computational model to generate both robotic-like and human-like spatiotemporal trajectories. As such, we added to iDeLog [10] a new module about generation of robotic movements.

These movements were synthetically designed by fitting a constant speed to each stroke between virtual target points. Next, the synthetic velocity profile was obtained by concatenating all these speeds. To this aim, a square- wave signal was obtained. According to our observations, the robotic velocity profile describes a trapezoidal pattern. In this work, this pattern was achieved by smoothing the square-wave signals with realistic aspect. Similar to the human-like case, the trajectories between target points were resampling with our trapezoidal-based velocity profiles.

III. GENERATING SPECIFIC KINEMATICS WITH INDUSTRIAL ROBOTS

The robot motion is presented as a sequence of *(x, y, z, t)* tuples representing the desired TCP frame position at the given time *t*. As the TCP frame orientation is not specified, we have fixed it to RPY(0, 3/4 π, 0), thus obtaining a completely defined pose for each waypoint. Since the robot control demands the task be specified in the joint space, for each of the poses along the trajectory we obtained an inverse kinematics solution, using an iterative Jacobian solver available from the RobWork library [18]. The Jacobian solver is well suited for such a task, since it tends to generate well-behaved continuous solution for a given path. We have thus obtained a sequence of ($q_1$, $q_2$, $q_3$, $q_4$, $q_5$, $q_6$, t) tuples specifying the desired trajectory in the joint space.

In our experiment, we have used a Universal Robots UR3 [19] manipulator mounted on a setup shown in Fig. 1. The robot arm has 500 mm maximum reach, with 0.1 mm repeatability and is able to lift a payload of up to 3 kg.

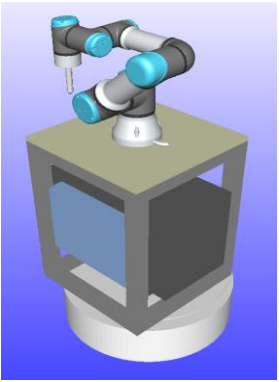
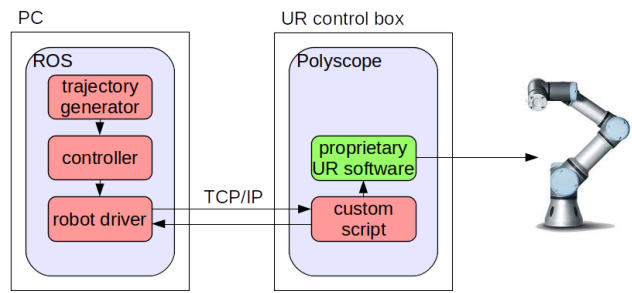

Fig.1 The UR3 setup at the BUT laboratory.

Fig. 2 The control scheme for the trajectory execution using the UR3 robot.

The UR3 robot can be controlled in several modes. Standard UR programming allows for the trajectory generation through the waypoint specification (either in joint space or in the Cartesian space). Since such approach is not well suited for precise speed control, we have opted for online joint velocity control using a scheme presented in Fig. 2.

In this scheme, we have implemented a ROS package [20] running on a PC workstation, including a trajectory generator, robot velocity controller and a robot driver. The trajectory generator generates trajectory positions and velocities (in joint space) for the current time using a 3rd degree spline interpolators. The robot velocity controller has a PI feed-forward structure (see Fig. 3) to compute the desired velocity of the robot joints.

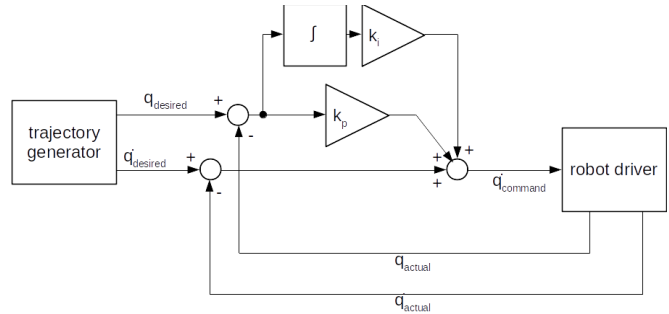

Fig. 3 Robot controller structure.

The robot driver communicates with the robot control box and sends encoded desired velocity commands over a TCP/IP connection. The robot control box is running a custom URScript program [21] to receive the commands and drive the joint motors. The feedback, i.e. the actual positions and velocities of the joints, is obtained through the UR data interface, running with the frequency of 125 Hz.

## IV. EXPERIMENTAL RESULT

We first describe the data developed in this work. Next, we guarantee the similarity between the generated movement with iDeLog3D and the output robot movement in terms of Signal-to-noise-ratio.

### A. Used dataset

For this paper, we designed ten movements with iDeLog. On the one hand, five random movements described a bell-shaped velocity profile as a sum of consecutive lognormal functions. On the other hand, other five movements were generated by describing a trapezoidal velocity profile.

To this aim, the movements were about four and five target points. Typically, an industrial robot is programed to make repetitive movements. Accordingly, our movements were repeated three times for the experiments.

They target points were initially programed to be executed in a tridimensional parallelepiped. To fit the trajectories movements into the UR3's workspace, they were finally moved, scaled and rotated without modifying the velocity shapes.

It is worth highlighting that both the robotic and bell-shaped velocities were generated keeping a similar duration between strokes. It allows to avoid bias in the performance for the duration of the movements.

### B. Similarities between input and output robotic movement

We have performed an experiment of executing a set of 10 trajectories (5 with uniform velocity profile and 5 with lognormal velocity profile) using an UR3 robot arm. The results are presented in the figures below. Fig. 4 shows the desired ($v_{PC}$), i.e. computer generated, and actual ($v_{UR3}$) robot TCP velocity during the typical uniform motion execution, while Fig. 5 show the desired computer generated, and actual robot TCP velocity during the typical lognormal trajectory execution. Since the velocities are computed as the TCP position derivative, smoothing was applied in the figures to reduce the actual velocity signal noise.

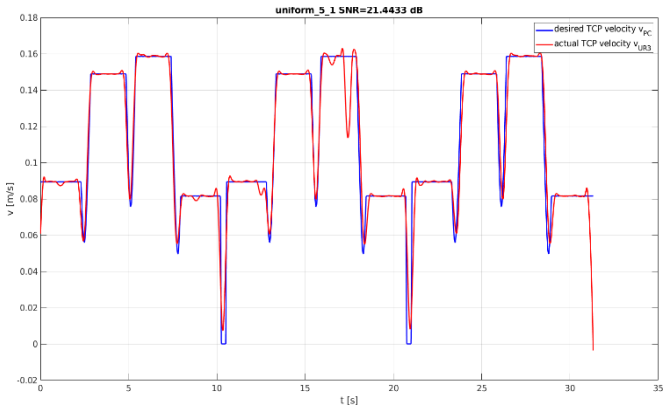

Fig. 4. Desired, i.e. computer generated, and actual robot TCP velocities during the execution of the uniform_5_1 motion profile

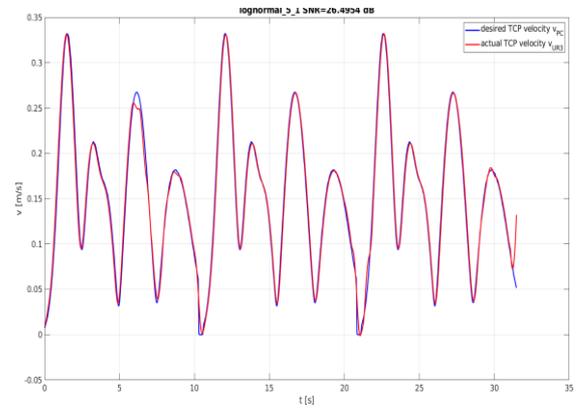

Fig. 5. Desired, i.e. computer generated, and actual robot TCP velocities during the execution of the lognormal_5_1 motion profile.

To assess the quality of the resulting trajectories we have computed the SNR (signal-to-noise-ratio) value of the TCP velocity for all the executed motions (see Table I). SNR is calculated as follows:

$$SNR_{dB} = 10\log_{10}\left(\frac{P(v_{PC})}{P(v_{UR3}-v_{PC})}\right) \quad (10)$$

where $P(v_{PC})$ is the power of the computer generated velocity signal, and the $P(v_{UR3} - v_{PC})$ is the power of the noise signal, taken as the difference between the actual velocity $v_{UR3}$ and the computer generated velocity $v_{PC}$.

TABLE I.

SNR BETWEEN SYNTHETIC AND REAL ROBOT MOVEMENTS

| uniform | | lognormal | |
|---|---|---|---|
| name | SNR [dB] | name | SNR [dB] |
| uniform_4_1 | 19.02 | lognorm_4_1 | 29.90 |
| uniform_4_2 | 17.58 | lognorm_4_2 | 15.96 |
| uniform_5_1 | 21.44 | lognorm_5_1 | 26.50 |
| uniform_5_2 | 18.92 | lognorm_5_2 | 16.56 |
| uniform_5_3 | 19.75 | lognorm_5_3 | 21.88 |

## V. CONCLUSIONS AND FUTURE WORK IDEAS

In this work we demonstrated the possibility to control the velocity of a specific industrial robot, UR3, to generate human-like or robotic-like movements. The movements were generated within a tridimensional parallelepiped as a set of random target points, which simulate a possibility trajectory to travel by an industrial robot when manufacturing. The spatiotemporal trajectories were obtained by using iDeLog, which is able to generate trajectories with both uniform *(robotic)* or bell-shaped *(human-like)* velocity profiles. As a proof of concepts, these trajectories were reproduced by a UR3 robot obtained challenging performance when the spatiotemporal trajectories of the robots were compared to the programmed with iDeLog.

In our future works we plan to extend this work to other industrial robots, like the ABB IRB-120 robot. We also plan to design more movements including context-task where a person has to interact with the robot. We expect to improve the acceptability to use industrial robots in HRI with human-like movements.

ACKNOWLEDGMENT

This research was funded by the Spanish government's PID2019-109099RB-C41 research projects, the European Union FEDER program/funds and by the original research work WZ/WM-IIM/1/2019 funded by the Polish Ministry of Science and Higher Education.